\documentclass[osajnl,twocolumn,showpacs,superscriptaddress,10pt]{revtex4-1} %% use 11pt for Applied Optics
\usepackage{amsmath}
\usepackage{amssymb}
\usepackage{graphicx}
\usepackage[ruled]{algorithm2e}
\usepackage{algorithmic}
\usepackage{subfig}
\usepackage{cases}
\usepackage{hyperref} %% optional

\begin{document}

\title{Fourier-Bessel rotational invariant eigenimages}

\author{Zhizhen Zhao}\email{Corresponding author: zhizhenz@princeton.edu}
\affiliation{Physics Department, Princeton University Jadwin Hall, Washington Road, Princeton, NJ. 08540, USA}

\author{Amit Singer}
\affiliation{Mathematics Department and PACM, Princeton University, Fine Hall, Washington Road, Princeton, NJ. 08540, USA}

\begin{abstract}We present an efficient and accurate algorithm for principal component analysis (PCA) of a
large set of two-dimensional images, and, for each image, the set of its uniform rotations in the plane and its reflection.
The algorithm starts by expanding each image, originally given on a Cartesian grid, in the Fourier-Bessel basis for the disk. Because the images are essentially bandlimited in the Fourier domain, we use a sampling criterion to truncate the Fourier-Bessel expansion such that the maximum amount of information is preserved without the effect of aliasing. The constructed covariance matrix is invariant to rotation and reflection and has a special block diagonal structure. PCA is efficiently done for each block separately. This Fourier-Bessel based PCA detects more meaningful eigenimages and has improved denoising capability compared to traditional PCA for a finite number of noisy images. \end{abstract}

\maketitle

\section{Introduction}
Principal component analysis (PCA) is a classical method for dimensionality reduction, compression and denoising. The principal components are the eigenvectors of the sample covariance matrix. In image analysis, the principal components are often referred to as ``eigenimages" and they form a basis adaptive to the image set. In some applications, including all planar rotations of the input images for PCA is advantageous. For example, in single particle reconstruction (SPR) using cryo-electron microscopy \cite{Frank}, the 3D structure of a molecule needs to be determined from many noisy 2D projection images taken at unknown viewing directions. PCA, known in this field as multivariate statistical analysis (MSA) is often a first step in SPR \cite{vanHeel}. Inclusion of the rotated images for PCA is desirable, because such images are just as likely to be obtained in the experiment, by in-plane rotating either the specimen or the detector. When all rotated images are included for PCA, then the eigenimages have a special separation of variables form in polar coordinates in terms of radial functions and angular Fourier modes \cite{Hilai,Perona,Uenohara}. It is easy to steer the eigenimages by a simple phase shift, hence the name ``steerable PCA". Computing the steerable PCA efficiently and accurately is however challenging.

The first challenge is to mitigate the increased computational cost associated with replicating each image multiple times. Efficient algorithms for steerable PCA were introduced in \cite{Jogan,Ponce} with computational complexity almost similar to that of traditional PCA on the original images without their rotations. Current efficient algorithms for steerable PCA first map the images from Cartesian grid to polar grid, using, e.g., interpolation. Because the transformation from Cartesian to polar is not unitary, the eigenimages corresponding to images mapped to polar grid are not equivalent to transforming the original eigenimages from Cartesian to polar.

The second challenge is associated with noise. The non-unitary transformation from Cartesian to polar changes the noise statistics. Interpolation transforms additive white noise to correlated (i.e., colored) noise. As a consequence, spurious eigenimages and eigenvalues corresponding to colored noise may arise in the eigen-analysis. The na\"ive algorithm for steerable PCA that replicates each image multiple times also mistreats the noise: While the realization of noise is independent between the original images, the realization of noise among duplicated images is dependent. This in turn can lead to noise-induced spurious eigenimages and to artifacts in the leading eigenimages.

We present a new efficient and accurate algorithm for computing the steerable PCA that meets these challenges. This is achieved by combining into the steerable PCA framework a sampling criterion similar to the criterion proposed by Klug and Crowther~\cite{Klug} in a different context of reconstructing a 2D image from its 1D projections. We represent the images in a truncated Fourier-Bessel basis, in which the number of radial components is decreasing with the angular frequency. Our sampling criterion implies that the covariance matrix of the images has a block diagonal structure where the block size decreases as a function of the angular frequency. The block diagonal structure of the covariance matrix was observed and utilized in previous works on steerable PCA. However, while in all existing methods for steerable PCA the block size is constant, here the block size shrinks with the angular frequency. The incorporation of the sampling criterion into the steerable PCA framework is the main contribution of this paper.

\section{Sampling criterion}

We assume that the set of images correspond to spatially limited objects. By appropriate scaling of the pixel size, we can assume that the images vanish outside a disk of radius $1$. The eigenfunctions of the Laplacian in the unit disk with vanishing Dirichlet boundary condition are the Fourier-Bessel functions. Hence, they form an orthogonal basis to the space of squared-integrable functions over the unit disk, and it is natural to expand the images in that basis.
%Because the Fourier-Bessel transform is unitary, additive white Gaussian noise does not change the statistics under such transform.
The Fourier-Bessel functions are given by
\begin{equation}
\label{eq:FB}
\psi^{k q}(r,\theta)=
\begin{cases}
N_{kq}J_k(R_{kq} r)e^{\imath k\theta} , & r \leq 1  \\
0 , & r > 1,
\end{cases}
\end{equation}
where $N_{kq}$ is a normalization factor; $J_k$ is the Bessel function of integer order $k$; and $R_{kq}$ is the $q^{\mathrm{th}}$ root of the equation
\begin{equation}
\label{eq:zeros}
J_k( R_{kq})= 0.
\end{equation}
The functions $\psi^{kq}$ are normalized to unity, that is
\begin{equation}
\int_0^{2\pi }\int_{0}^{1} \psi^{kq} (\psi^{kq})^*r \,dr \,d \theta = 1.
\end{equation}
The normalization factors are given by
\begin{equation}
\label{eq:norm}
N_{k,q}= \frac{1}{\sqrt{\pi} | J_{k+1}( R_{kq})|}.
\end{equation}

% This part is added to show why Bessel zeros are important:
We use here the following convention for the 2D Fourier transform of a function $f$ in polar coordinates
\begin{equation}
\mathcal{F}(f)(k_0,\phi_0) = \int_0^{2\pi} \int_0^\infty f(r,\theta) e^{-2\pi \imath k_0 r \cos (\theta-\phi_0)} r \,dr\,d\theta.
\end{equation}
The 2D Fourier transform of the Fourier-Bessel functions, denoted $\mathcal{F}(\psi^{kq})$, is given in polar coordinates as
\begin{equation}
\label{eq:FT_FB}
\mathcal{F}(\psi^{k q})(k_0, \phi_0) = 2 \sqrt{\pi} (-1)^q (-\imath)^k R_{kq} \frac{J_k (2\pi k_0)}{(2\pi k_0)^2-R_{kq}^2} e^{\imath k \phi_0}.
\end{equation}
This result is typically derived using the Jacobi-Anger identity
\begin{equation}
e^{\imath z\cos \theta} =\sum_{n=-\infty}^{\infty} \imath^n J_n(z) e^{\imath n \theta}.
\end{equation}

Notice that the Fourier transform $\mathcal{F}(\psi^{k q})(k_0, \phi_0)$ vanishes on concentric rings of radii $k_0 = \frac{R_{kq'}}{2\pi}$ with $q'\neq q$. The maximum of $|\mathcal{F}(\psi^{k q})(k_0, \phi_0)|$ is obtained near the ring $k_0 = \frac{R_{kq}}{2\pi}$, as can be verified from the asymptotic behavior of the Bessel functions (cf. eq. (\ref{asym})). For images that are sampled on a squared Cartesian grid of size $2L \times 2L$ pixels, with grid size $1/L$ (corresponding the square $[-1,1]\times [-1,1]$) the sampling rate is $L$ and the corresponding Nyquist frequency (the bandlimit) is $\frac{L}{2}$. Due to the Nyquist criterion, the Fourier-Bessel expansion requires components for which
\begin{equation}
\label{criterion}
\frac{R_{kq}}{2\pi} \leq \frac{L}{2},
\end{equation}
because other components represent features beyond the resolution and their inclusion would result in aliasing.
The asymptotic behavior of the Bessel functions
\begin{equation}
\label{asym}
J_k ( R_{kq} r) \sim \sqrt{\frac{2}{\pi R_{kq} r}}\,\cos ( R_{kq} r -\frac{k \pi}{2} - \frac{\pi}{4})
\end{equation}
for $R_{k q}r\gg | k^2 - \frac{1}{4}|$, suggests that the roots are asymptotically
\begin{equation}
\label{Rkq}
R_{kq} \sim \frac{\pi}{2}(k+2q-\frac{1}{2}),
\end{equation}
which is indeed the first term of the asymptotic expansion for the roots (see \cite{McMahon}). Eqs. (\ref{criterion}) and (\ref{Rkq}) lead to the sampling criterion
\begin{equation}
\label{sampling}
k + 2q \leq 2L + \frac{1}{2}.
\end{equation}
In practice, we do not rely on the asymptotic formula (\ref{Rkq}). Instead, we find the roots of the Bessel functions numerically, and check directly which components satisfy the criterion $R_{kq} \leq \pi L$. The number of components satisfying $|k|+2q \leq 2L + \frac{1}{2}$ is approximately $2L^2$, which is smaller than $\pi L^2$ (the number of pixels inside the unit disk) by a factor of $\frac{2}{\pi}$. We remark that the cut-off criterion (\ref{sampling}) is similar to the criterion in \cite{Klug} but is not identical to it. Specifically, the criterion in \cite{Klug} has a different cut-off value for $k+2q$. The difference stems from the fact that \cite{Klug} studies a different problem, namely, the 2D reconstruction problem of an image from its 1D line projections.

\section{Fourier-Bessel expansion of images sampled on a Cartesian grid}

\begin{figure}
\begin{center}
\subfloat[$L=6$]{
\includegraphics[width=0.4\columnwidth]{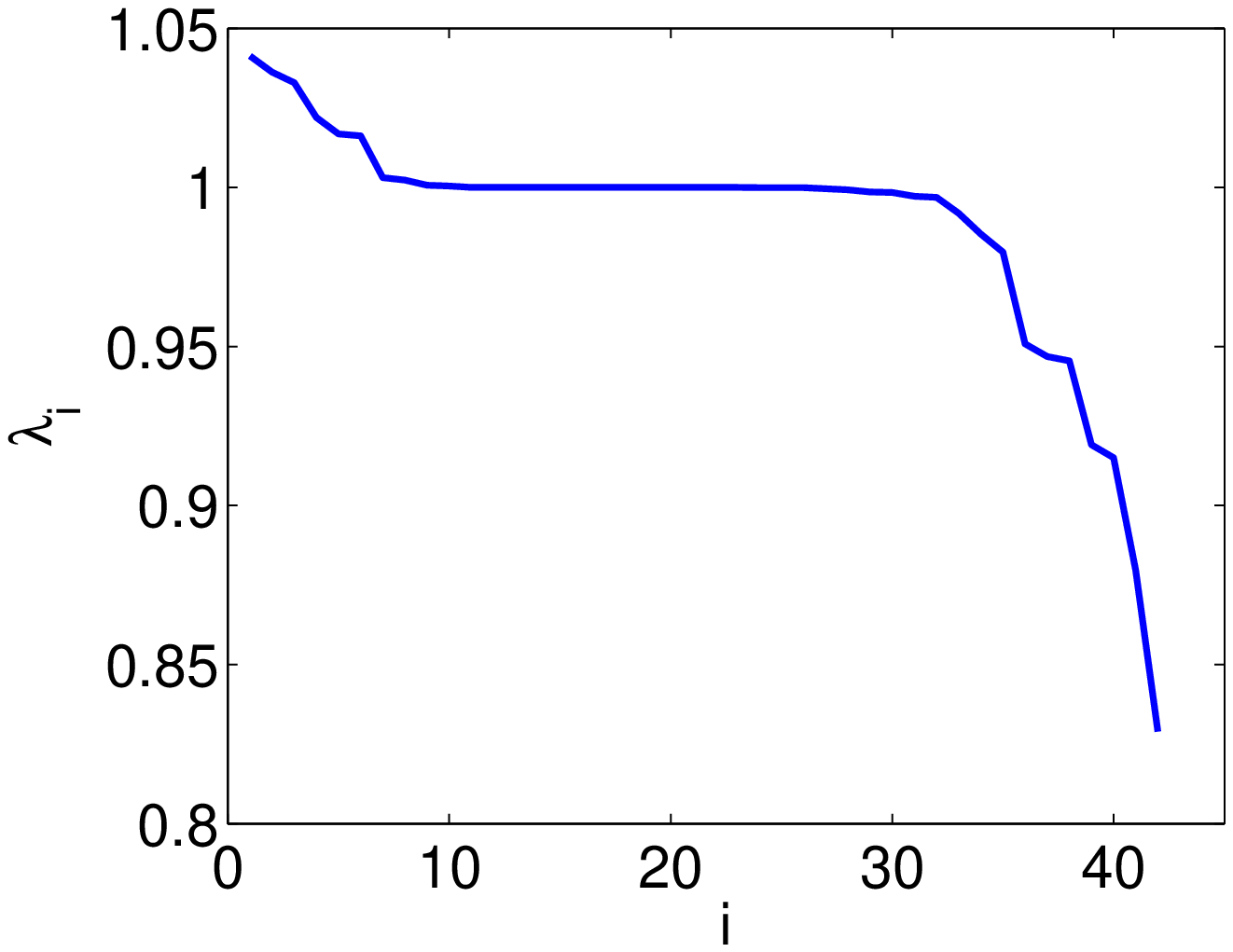}
}
\subfloat[$L=60$]{
\includegraphics[width=0.4\columnwidth]{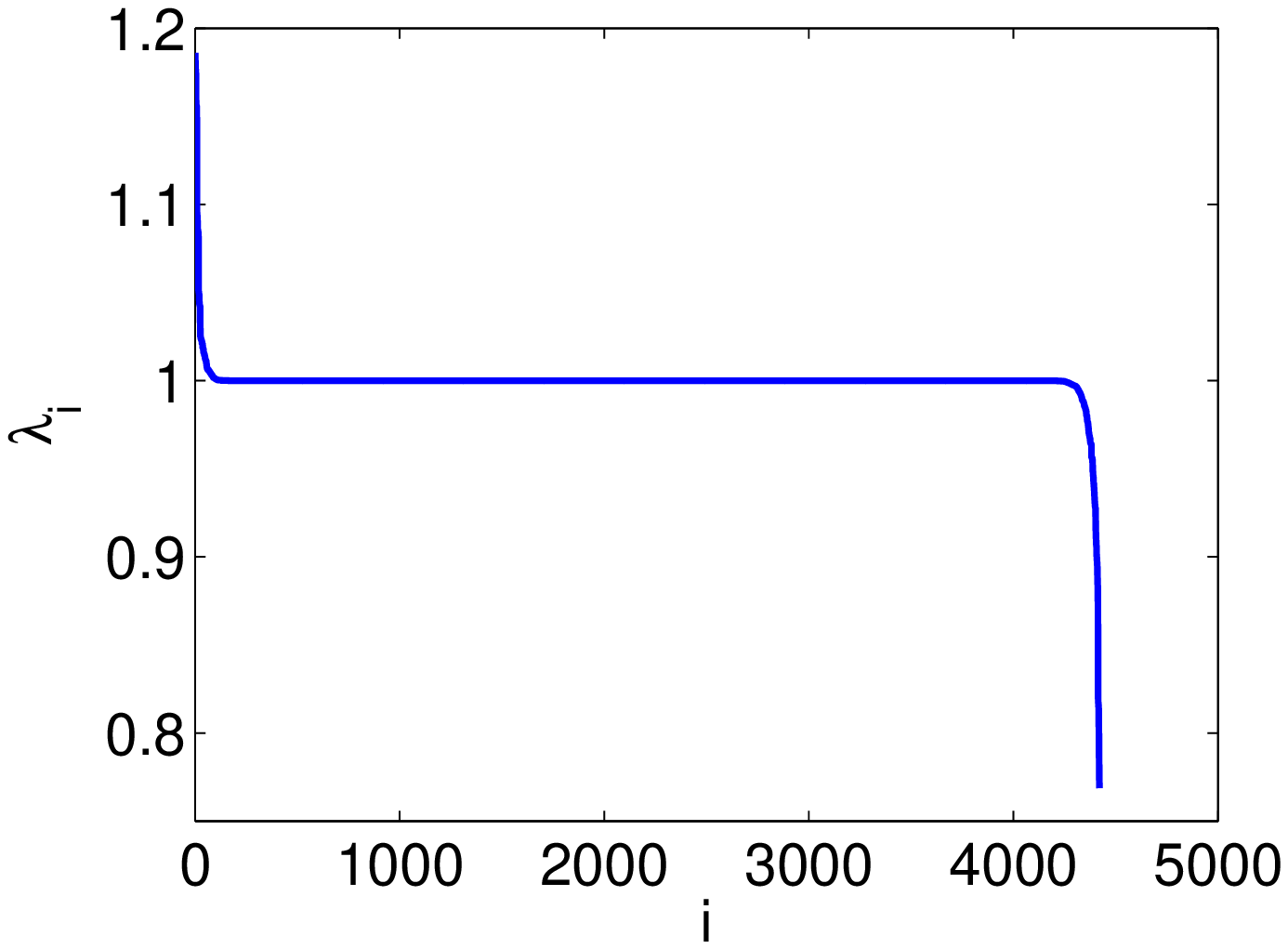}
}
\end{center}
\caption{Eigenvalues of $(\Psi^\dagger \Psi)^{-1}$, where $\Psi$ is the truncated Fourier-Bessel transform for $L = 6$ and $L=60$ pixels respectively. This is also the spectrum of transformed white noise images (see text). The eigenvalues are close to 1, indicating that the discretized Fourier-Bessel transform is approximately unitary and that white noise remains approximately white. For $L = 60$, among the eigenvalues, only 18 are above $1.05$ and 34 are below $0.95$. For $L = 6$, among the eigenvalues, there are 6 below 0.95. }
\label{fig:Eig_A}
\end{figure}

Suppose $I_1,\ldots,I_n$ are $n$ images sampled on a Cartesian grid. We denote by $\tilde{I}_i$ the continuous approximation of the $i$'th image in terms of a truncated Fourier-Bessel expansion including only components satisfying the sampling criterion, namely
\begin{equation}
\label{lsqr}
\tilde{I}_i(r, \theta) = \sum_{k=-2L}^{2L} \sum_{q=1}^{p_k}  a^i_{k,q} \psi^{kq} (r, \theta),
\end{equation}
where for each $k$, $p_k$ denotes the number of components satisfying $R_{kq}\leq \pi L$.
We choose the expansion coefficients $a^i_{k,q}$ that minimize the squared distance between $I_i$ and $\tilde{I}_i$, with the latter restricted to the grid points, that is,
$a^i_{k,q}$ are the least squares solution to the overdetermined linear system obtained from eq. (\ref{lsqr}) by evaluating the images and the Fourier-Bessel functions at the grid points. More specifically, let $\Psi$ be the matrix whose entries are evaluations of the Fourier-Bessel functions at the grid points, with rows indexed by the grid points and columns indexed by angular and radial frequencies. Then, the coefficient vector ${a}^i$ of the $i$'th image is the solution of $\min_{a^i} \|\Psi a^i - I_i\|^2$, given by
\begin{equation}
\label{eq:a}
a^i = (\Psi^\dagger \Psi)^{-1}\Psi^\dagger I_i.
\end{equation}
The orthogonality of the Fourier-Bessel functions on the disk does not necessarily imply that their discretized counterparts are orthogonal. That is, the matrix $\Psi^\dagger \Psi$ may differ from the identity matrix. The columns of the matrix $\Psi$ are guaranteed to approach orthogonality only as the grid is indefinitely refined  (e.g, as $L\to \infty$). In practice, we have numerically observed that the columns of $\Psi$ are approximately orthogonal already for moderate values of $L$. In particular, white noise remains approximately white after the transformation, since the eigenvalues of $(\Psi^\dagger \Psi)^{-1}$ are close to 1 (see Figure \ref{fig:Eig_A}). Indeed, for white noise images, the mean and covariance satisfy $\mathbb{E}[I_i]=0$ and $\mathbb{E}[I_iI_i^\dagger]=\sigma^2 \text{I}$, where $\sigma^2$ is the noise variance and $\text{I}$ is the identity matrix. Therefore, eq. (\ref{eq:a}) and the linearity of expectation yield $\mathbb{E}[a^i] =0$ and $\mathbb{E}[a^i{a^i}^\dagger] = \sigma^2 (\Psi^\dagger \Psi)^{-1}$.

In the previous work~\cite{Ponce} on fast computation of steerable principal components for large data set, images are re-sampled on a polar grid by the polar Fourier transform. %Eigenimages are the 2D inverse Fourier transform of the eigenvectors of the sample covariance matrix of polar Fourier transformed images.
The polar Fourier transform is non-unitary and therefore causes artifacts in the eigenimages. Suppose the linear transformation is described by the matrix $A$. The sample covariance matrix $C_A$ of the transformed images is related to the sample covariance matrix $C$ of the original images by
\begin{equation}
C_A = A C A^\dagger.
\end{equation}
Unless $A$ is unitary, there is no simple way to transform the eigenvectors of $ACA^\dagger$ to the eigenvectors of $C$. The advantage of using the Fourier-Bessel transform with the sampling criterion that adapts to the band limit of the images is that such transform is approximately unitary (Fig. \ref{fig:Eig_A}).

\section{The sample covariance matrix}

It is easy to ``steer" the continuous approximation $\tilde{I}_i$ of $I_i$ (eq. (\ref{lsqr})).
If $\tilde{I}_i^\alpha$ denotes the rotation of the $i$'th image by an angle $\alpha$, then
\begin{equation}
\label{eq:Ii}
\tilde{I}_i^\alpha (r, \theta)= \tilde{I}_i(r, \theta-\alpha)= \sum_{k, q} a^i_{k,q} e^{-\imath k \alpha} \psi^{kq} (r, \theta).
\end{equation}
Because $J_{-k}(x) = (-1)^k J_k(x)$, for real valued images $a_{-k,q} = (-1)^k a_{k,q}^*$. Also, under reflection $a^i_{k,q}$ changes to $a^i_{-k, q}$, and the reflected image, denoted $\tilde{I}_i^r$, is given by
\begin{eqnarray}
\tilde{I}_i^r(r, \theta) &=& \tilde{I}_i(r, \pi - \theta) = \sum_{k, q} a^i_{k,q} \psi^{kq} (r, \pi-\theta) \\
&=& \sum_{k, q} a^i_{-k, q} \psi^{kq} (r, \theta).\label{eq:Ir}
\end{eqnarray}

The sample mean, denoted $\tilde{I}_{mean}$, is the continuous image obtained by averaging the continuous images and all their possible rotations and reflections:
\begin{equation}
\label{eq:mean}
\tilde{I}_{mean}(r,\theta) = \frac{1}{2n}\sum_{i=1}^n \frac{1}{2\pi} \int_0^{2\pi} \left[\tilde{I}_i^\alpha(r,\theta) + \tilde{I}_i^{\alpha, r}(r,\theta)\right] d\alpha
\end{equation}
Substituting eqs. (\ref{eq:Ii}) and (\ref{eq:Ir}) into (\ref{eq:mean}) we obtain
\begin{equation}
\label{eq:mean2}
\tilde{I}_{mean}(r,\theta) = \sum_{q=1}^{ p_0 } \left(\frac{1}{n}\sum_{i=1}^n a_{0,q}^i\right) \psi^{0q}(r,\theta).
\end{equation}
As expected, the sample mean image is radially symmetric, because $\psi^{0q}$ is only a function of $r$ but not of $\theta$.

\begin{figure}
\begin{center}
\includegraphics[width=0.6\columnwidth]{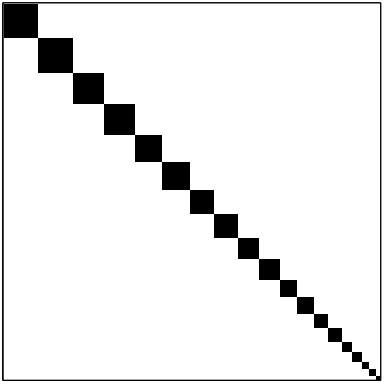}%
\end{center}
\caption{Illustration of the block diagonal structure of the rotational invariant covariance matrix. The block size shrinks as the angular frequency $k$ increases.}
\label{fig:Block}
\end{figure}

The sample covariance matrix built from the continuous images and all possible rotations and reflections is formally
\begin{eqnarray}
C &=& \frac{1}{2n} \sum_{i=1}^{n} \frac{1}{2\pi}\int_0^{2\pi}  \left[(\tilde{I}_i^\alpha - \tilde{I}_{mean}) (\tilde{I}_i^\alpha - \tilde{I}_{mean})^\dagger \right.\nonumber \\ &&\quad + \left.(\tilde{I}_i^{\alpha, r} - \tilde{I}_{mean})(\tilde{I}_i^{\alpha, r} - \tilde{I}_{mean})^\dagger)\right] d\alpha.
\end{eqnarray}
When written in terms of the Fourier-Bessel basis, the covariance matrix $C$ can be directly computed from the expansion coefficients $a^i_{k, q}$. Subtracting the mean image is equivalent to subtracting the coefficients $a^i_{0, q}$ with $\frac{1}{n}\sum_{i=1}^n a_{0,q}^i $, while keeping other coefficients unchanged. That is, we update
\begin{equation}
a^i_{0, q} \gets a^i_{0, q} - \frac{1}{n}\sum_{i=1}^n a_{0,q}^i.
\end{equation}
In the Fourier-Bessel basis the covariance matrix $C$ is given by
\begin{eqnarray}
& & C_{(k, q), (k', q')} \\
& = & \frac{1}{4\pi n} \sum_{i=1}^{n} \int_0^{2\pi} \left[a^i_{k, q} (a^i_{k', q'})^*  +  a^i_{-k, q} (a^i_{- k', q'})^*\right] e^{-\imath (k-k')\alpha } d \alpha \nonumber \\
& = & \delta_{k,k'}\frac{1}{n} \sum_{i=1}^{n}\mathfrak{R}\{a^i_{k, q } (a^i_{k, q'})^*\}.
\end{eqnarray}
The non-zero entries of $C$ correspond to $k=k'$, rendering its block diagonal structure (see Fig.\ref{fig:Block}).
Also, it suffices to consider $k\geq 0$, because $C_{(k,q),(k,q')} = C_{(-k,q),(-k,q')}$.
Thus, the covariance matrix can be written as the direct sum
\begin{equation}
C = \bigoplus_{k=0}^{2L  } C^{(k)}
\end{equation}
where $C^{(k)}_{q,q'} = \frac{1}{n} \sum_{i=1}^n \mathfrak{R}\{a^i_{k, q} (a^i_{k, q'})^*\}$ is by itself a sample covariance matrix of size $p_k \times p_k$. The block size $p_k$ decreases as the angular frequency $k$ increases (see Figure \ref{fig:Block}). We remark that the block structure of the covariance matrix in steerable PCA is well known; however, in previous works the block size is constant (i.e., independent of the angular frequency $k$). Our main observation is that the block size must reduce as the angular frequency increases in order to avoid aliasing.
Moreover, if the images are corrupted by independent additive white (uncorrelated) noise, then each block $C^{(k)}$ is also affected by independent additive white noise, because the Fourier-Bessel transform is unitary (up to grid discretization).

\section{Algorithm and computational complexity}
We refer to the resulting algorithm as Fourier-Bessel steerable PCA (FBsPCA). The steps of FBsPCA are summarized in Algorithm \ref{alg:FB_SPCA}.
\begin{algorithm}
\caption {Fourier-Bessel steerable PCA (FBsPCA)}
\begin{algorithmic}[1]
\REQUIRE $n$ pre-whitened images $I_1,\ldots,I_n$ sampled on a Cartesian grid of size $2L\times 2L$. \\
\STATE (Precomputation) Compute $\psi^{kq}$ for all $R_{kq}\leq \pi L$. \\
\STATE (Precomputation) Compute the Moore-Penrose pseudoinverse of $\Psi$. \\
\STATE For each $I_i$ compute $a^i_{k, q}$ using eq.~(\ref{eq:a}).\\
\STATE Estimate the $\tilde{I}_{mean}$ using eq.~(\ref{eq:mean2}). Subtract the mean image by changing $a^i_{0, q}$ to $a^i_{0, q}-\frac{1}{n}\sum_i a^i_{0, q}$.
\STATE For each $k = 0, 1, \dots, 2L$ compute the covariance matrix $C^{(k)}$ of size $p_k \times p_k$.\\
\STATE For each $k$ compute the eigenvalues of $C^{(k)}$, in descending order $\lambda_k^1 \ge \lambda_k^2 \cdots \geq \lambda^{p_k}_k$ and the associated eigenvectors $h_k^1, h_k^2,\ldots, h_k^{p_k}$.\\
\STATE The eigenimages $V^{kl}$ ($0\leq k \leq 2L$, $1\leq l \leq p_k$) are linear combinations of the Fourier-Bessel functions, i.e., $V^{kl} = \Psi^{(k)} h_k^l$, where $\Psi^{(k)} = [\psi^{k1}, \dots, \psi^{kp_k}]$.
\STATE (optional) Use an algorithm such as ~\cite{Kritchman} to estimate the number of components to choose for each $C^{(k)}$.
\end{algorithmic}
\label{alg:FB_SPCA}
\end{algorithm}
%Computational Complexity.
The computational complexity of FBsPCA (excluding pre-computation) is $O(n L^4+L^5)$, whereas the computational complexity of the traditional PCA (applied on the original images without their rotational copies) is $O(n L^4+L^6)$. The different steps of FBsPCA have the following computational cost. The cost for precomputing the Fourier-Bessel functions on the discrete Cartesian grid is $O(L^4)$, because the number of basis functions satisfying the sampling criterion is $O(L^2)$ and the number of grid points is also $O(L^2)$. The complexity of computing the pseudoinverse of $\Psi^\dagger \Psi$ is dominated by computing the singular value decomposition (SVD) of $\Psi$ in $O(L^6)$. Computing the psuedoinverse of $\Psi^\dagger \Psi$ is a precomputation that does not depend on the images. Alternatively, since the eigenvalues of $\Psi^\dagger \Psi$ are almost equal to one it can be approximated by the identity matrix. It takes $O(n L^4 )$ to compute the expansion coefficients $a^{i}_{k, q}$. The computational complexity of constructing the block diagonal covariance matrix is $O( n L^3 )$. Since the covariance matrix has a special block diagonal structure, its eigen-decomposition takes $O( L^4 )$. Constructing the steerable basis takes $O ( L^5 ) $. Therefore, the total computational complexity for FBsPCA without the precomputation is $O(n L^4 + L^5)$. For traditional PCA, computing the covariance matrix takes $O(n L^4 )$ and its eigen-decomposition takes $O( L^6 )$. Overall, the computational cost of FBsPCA is lower than that of the traditional rotational variant PCA.

\section{Results}
\begin{figure}[h!]
\begin{center}
\subfloat[Clean]{
\includegraphics[width=0.22\columnwidth]{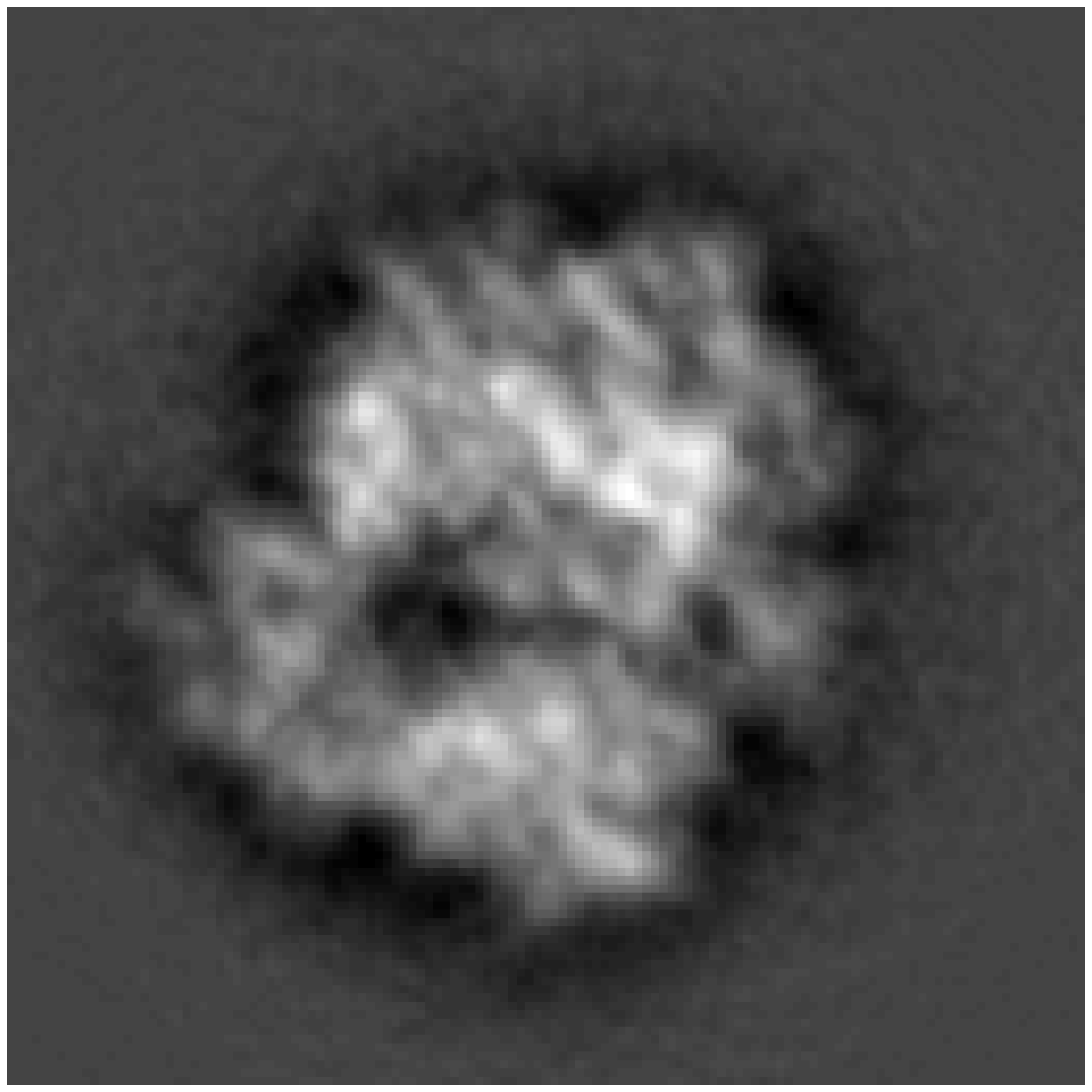}%
\label{fig:NMP0}
}\quad
\subfloat[SNR$=\frac{1}{10}$]{
\includegraphics[width=0.22\columnwidth]{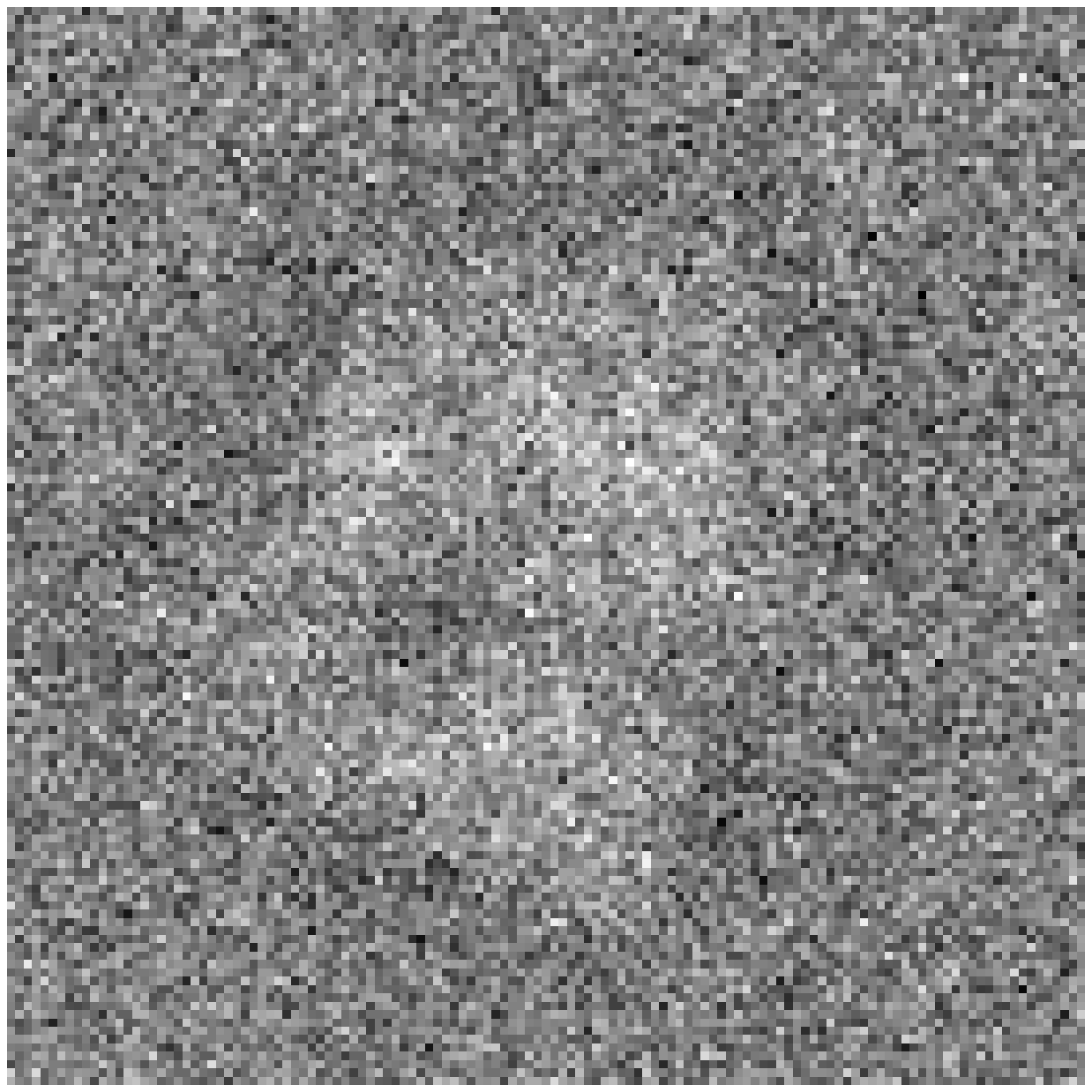}%
\label{fig:NMP1}
}\quad
\subfloat[SNR$=\frac{1}{50}$ ]{
\includegraphics[width=0.22\columnwidth]{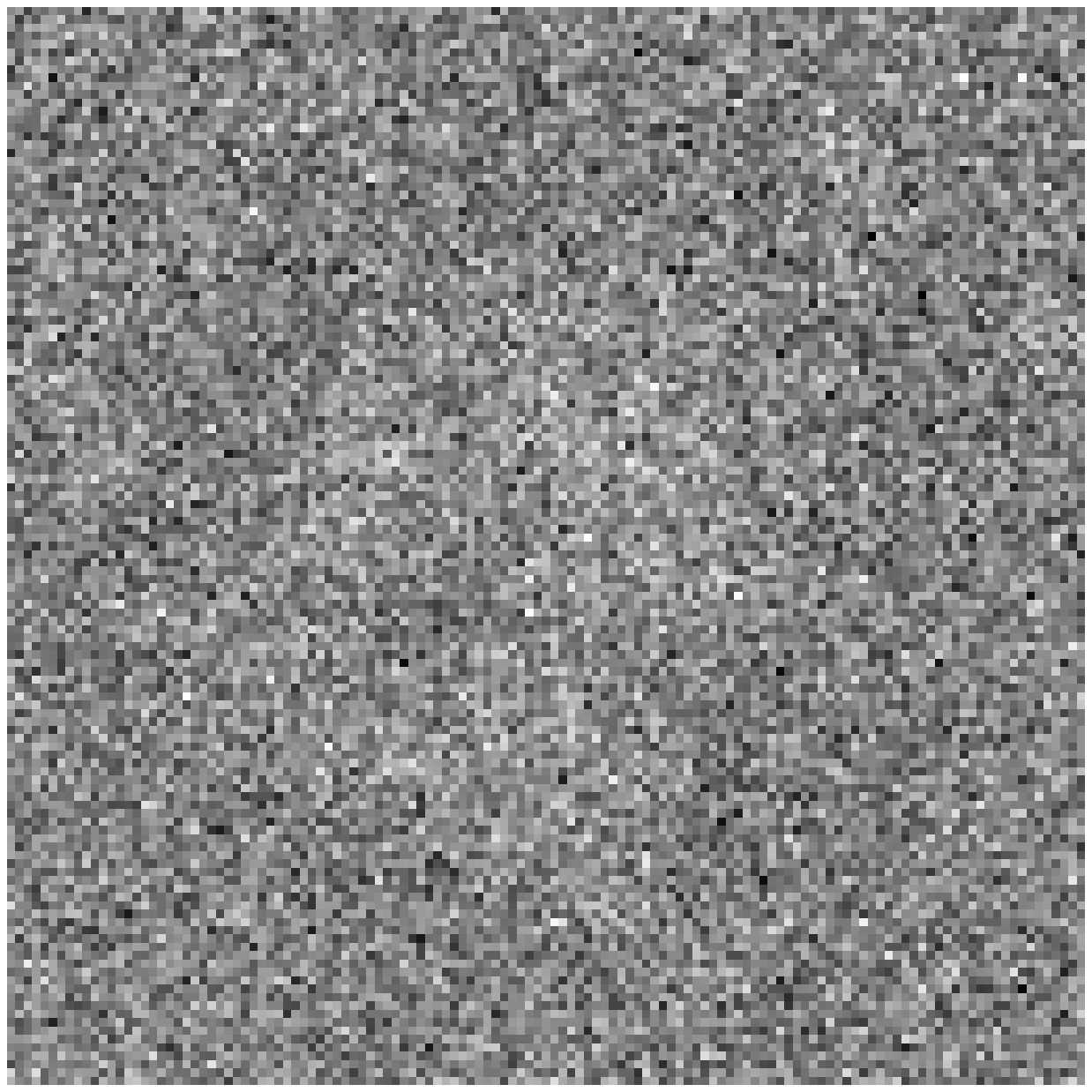}%
\label{fig:NMP20}
}
\end{center}
\caption{Simulated 70S ribosome projection images with different signal to noise ratio.}
\label{fig:Simulation}
\end{figure}

\begin{figure}[htb]
\begin{center}
\includegraphics[width=0.9\columnwidth]{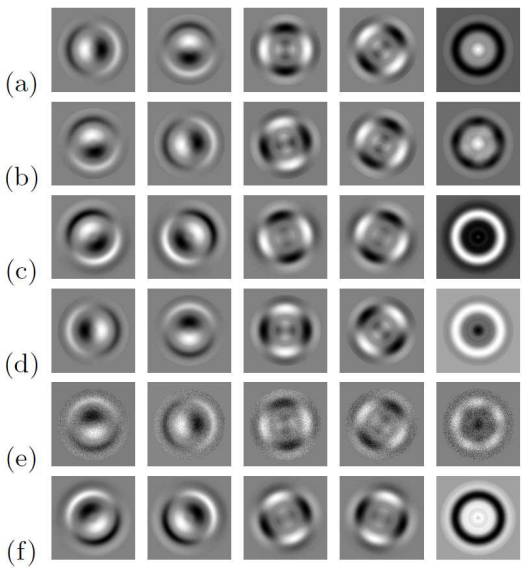}
\end{center}
\caption{Eigenimages for $10^4$ simulated $70S$ ribosome projection images. Clean images: (a) FBsPCA, (b) traditional PCA and (c) PTsPCA. Noisy images with SNR$=\frac{1}{50}$: (d) FBsPCA, (e) traditional PCA and (f) PTsPCA. Image size is $129 \times 129$ pixels and $L=55$ pixels.}
\label{fig:Bessel_basis}
\end{figure}

In the first experiment, we simulated $n=10^4$ clean projection images of {\it E. coli} 70S ribosome. The images are of size $129\times 129$ pixels, but the molecule is confined to a disk of radius $L=55$ pixels. We corrupted the clean images with additive white Gaussian noise at different levels of signal-to-noise ratio (SNR) (Fig. \ref{fig:Simulation}).
The running time of traditional PCA was $695$ seconds, compared to $85$ seconds (including precomputation) for FBsPCA, both implemented in MATLAB on a machine with 2 Intel(R) Xeon(R) CPUs X5570, each with 4 cores, running at 2.93 GHz.
The top $5$ eigenimages for noisy images agree with the eigenimages from clean projection images (Fig. \ref{fig:Bessel_basis}(a) and Fig. \ref{fig:Bessel_basis}(d)). The eigenimages generated by FBsPCA are much cleaner than the eigenimages from the traditional PCA (Fig. \ref{fig:Bessel_basis}(d) and Fig. \ref{fig:Bessel_basis}(e)). We also see that the eigenimages generated by steerable PCA with polar transform (PTsPCA) are consistent with the eigenvectors from the traditional PCA (see Fig. \ref{fig:Bessel_basis}(c) and Fig. \ref{fig:Bessel_basis}(f).

\begin{figure}[htb]
\begin{center}
\subfloat[$k=0$]{
\includegraphics[width=0.3\columnwidth]{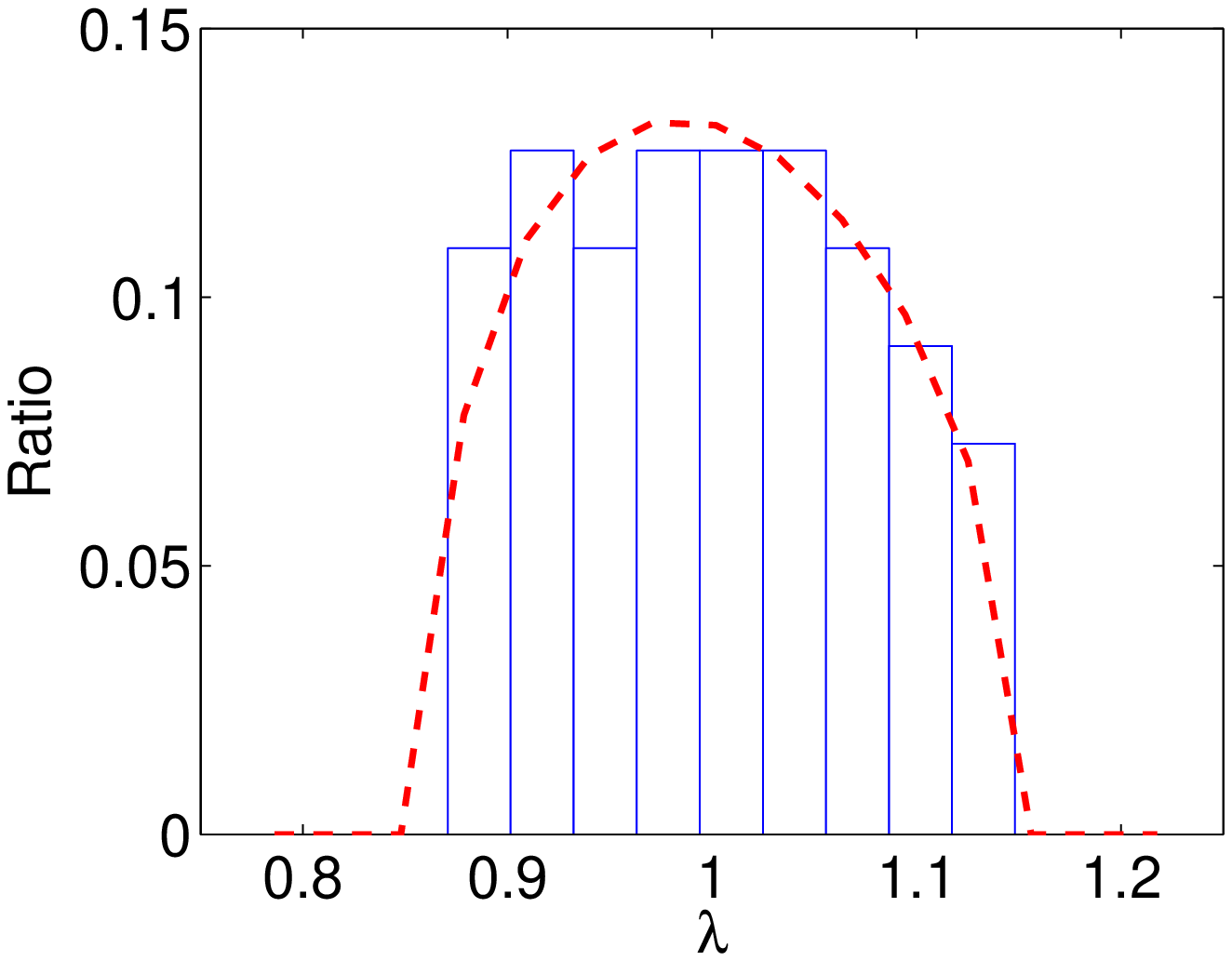}%
\label{fig:NMP0}
}
\subfloat[$k=1$]{
\includegraphics[width=0.3\columnwidth]{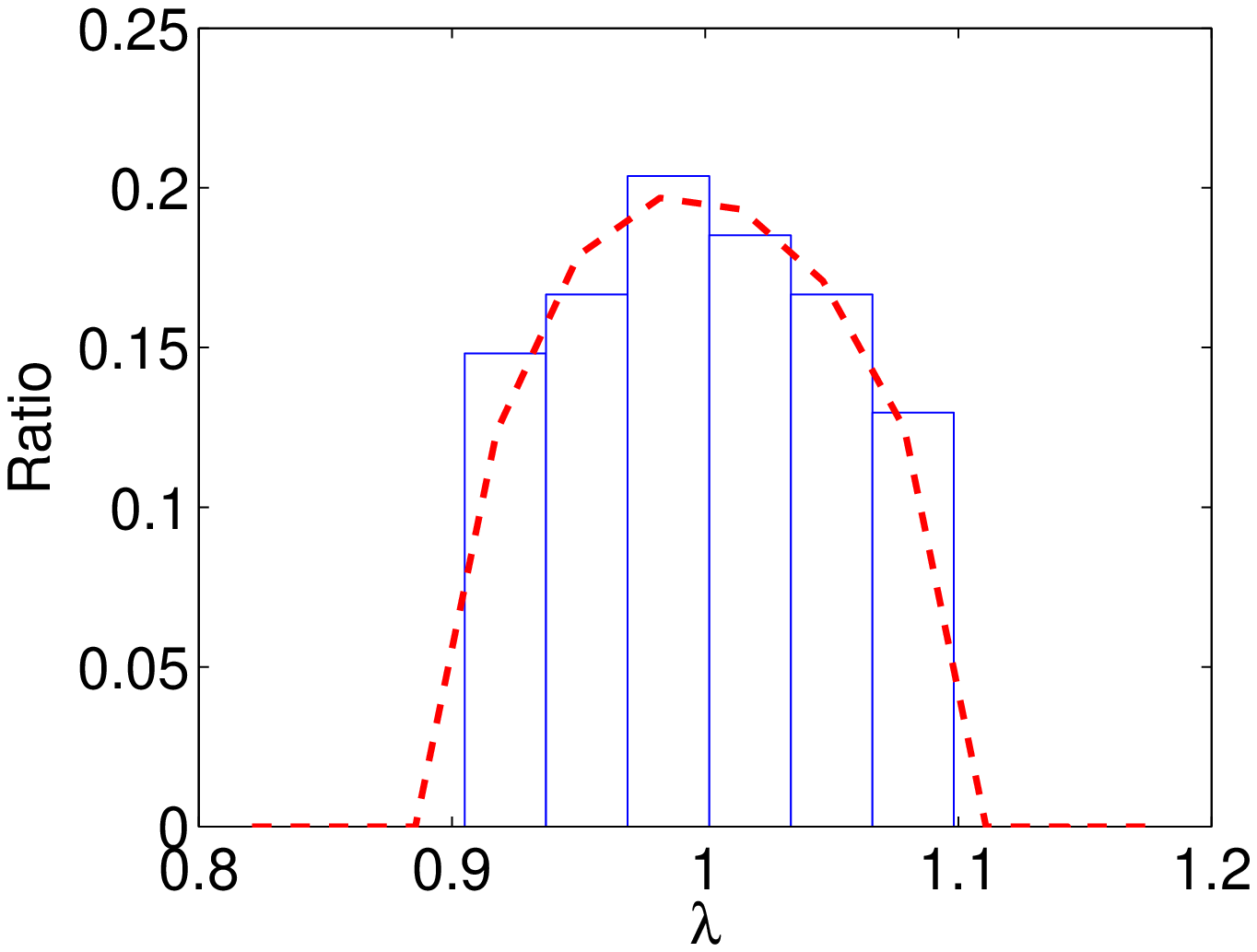}%
\label{fig:NMP1}
}
\subfloat[$k=20$ ]{
\includegraphics[width=0.3\columnwidth]{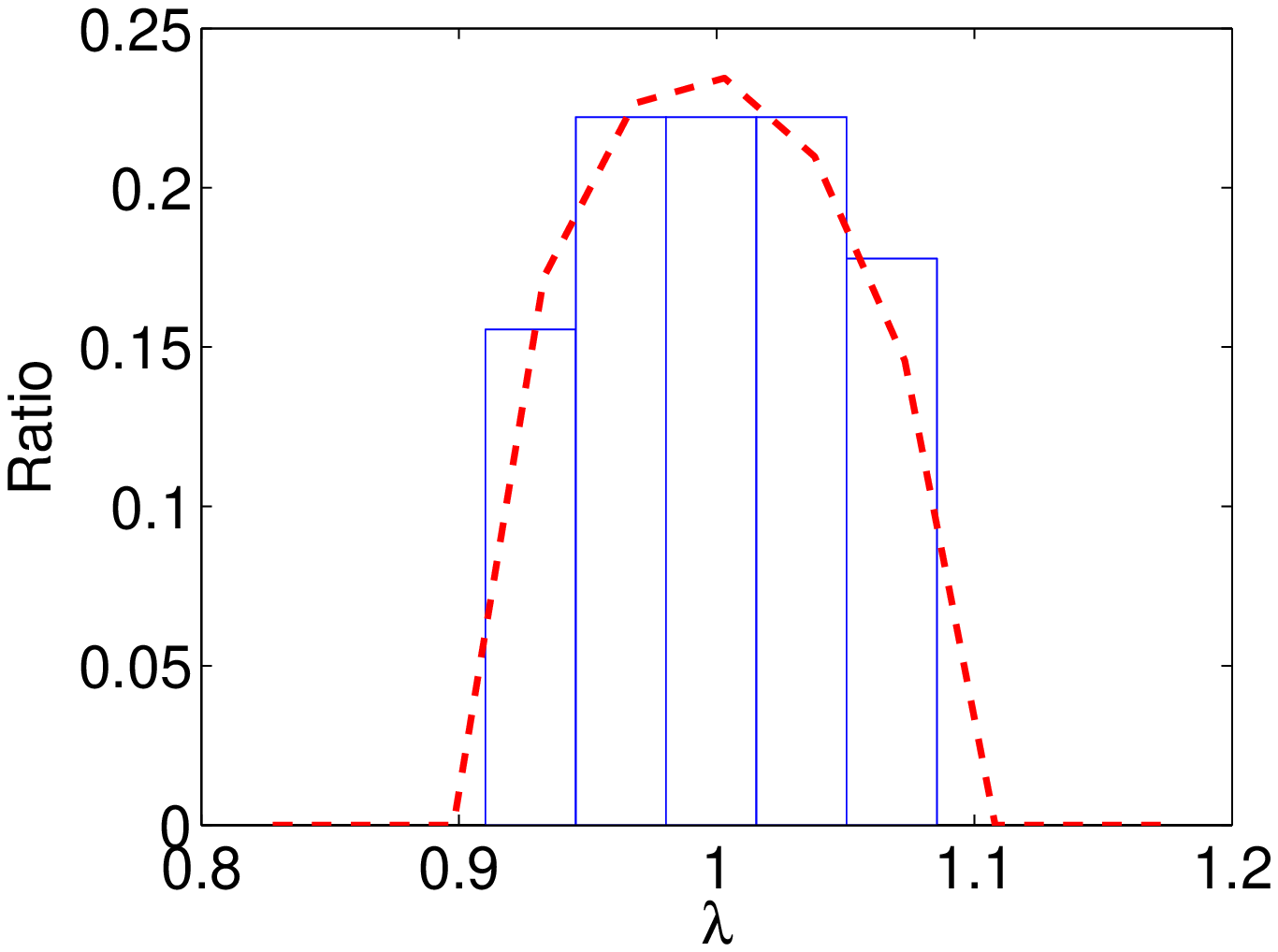}%
\label{fig:NMP20}
}
\end{center}
\caption{Histogram of eigenvalues of $C^{(k)}$ (with $k=0,1,20$) for images consisting of white Gaussian noise with mean $0$ and variance $1$. The dashed lines correspond to the Mar\v{c}enko-Pastur distribution. $n=10^4$, $L=55$ pixels and $\sigma^2 = 1$. }
\label{fig:Noise_MP}
\end{figure}

In another experiment, we simulated images consisting entirely of white Gaussian noise with mean $0$ and variance $1$ and computed the Fourier-Bessel expansion coefficients. The distribution of the eigenvalues for the rotational invariant covariance matrix with different angular frequencies $k$ are well predicted by the Mar\v{c}enko-Pastur distribution (see Fig. \ref{fig:Noise_MP}). This property allows us to use the method described in~\cite{Kritchman} to estimate the noise variance and the number of principal components to choose.

\begin{figure}
\begin{center}
\subfloat[$k=0$]{
\includegraphics[width=0.43\columnwidth]{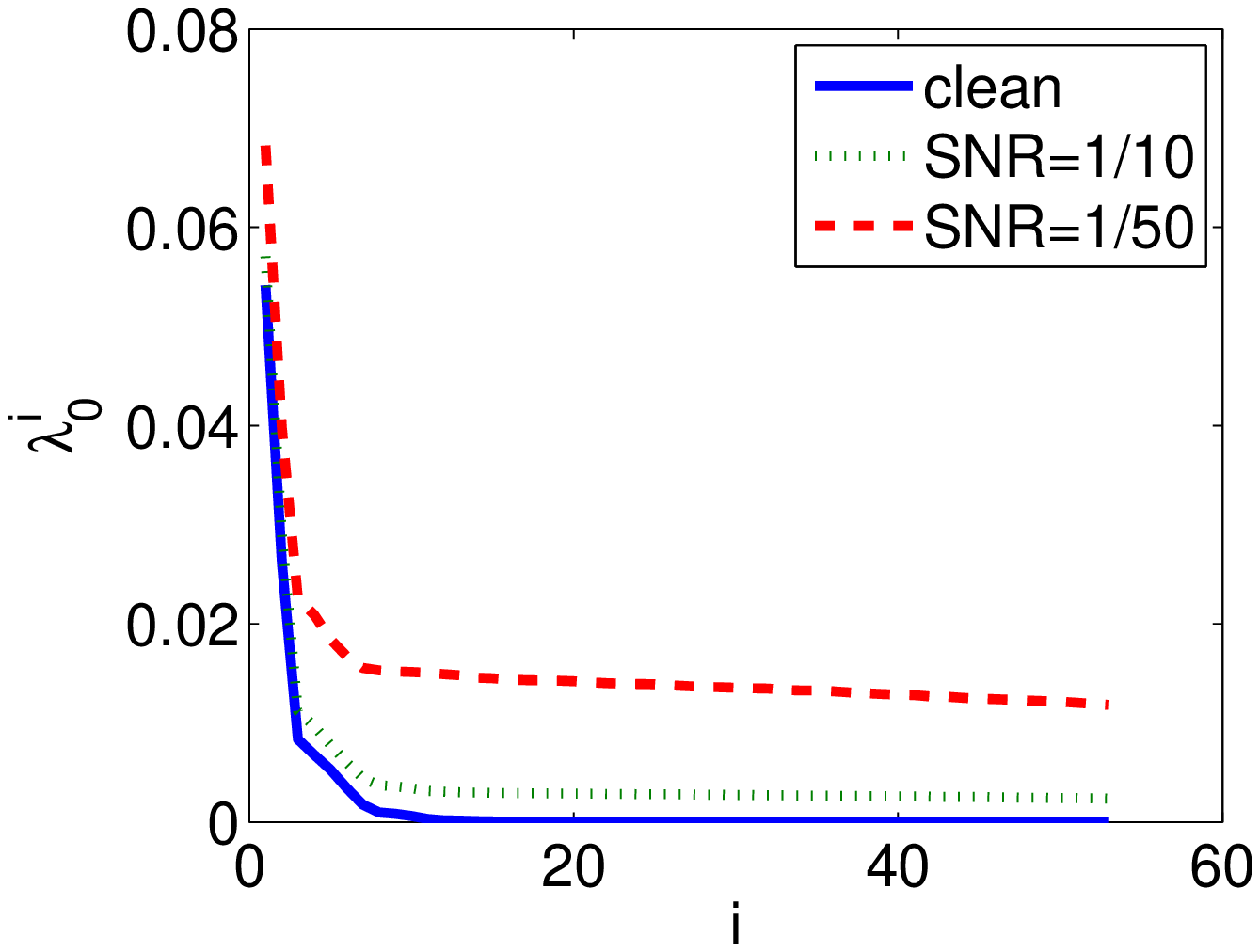}%
\label{fig:eigVal_k0}
}
\subfloat[$k=10$]{
\includegraphics[width=0.43\columnwidth]{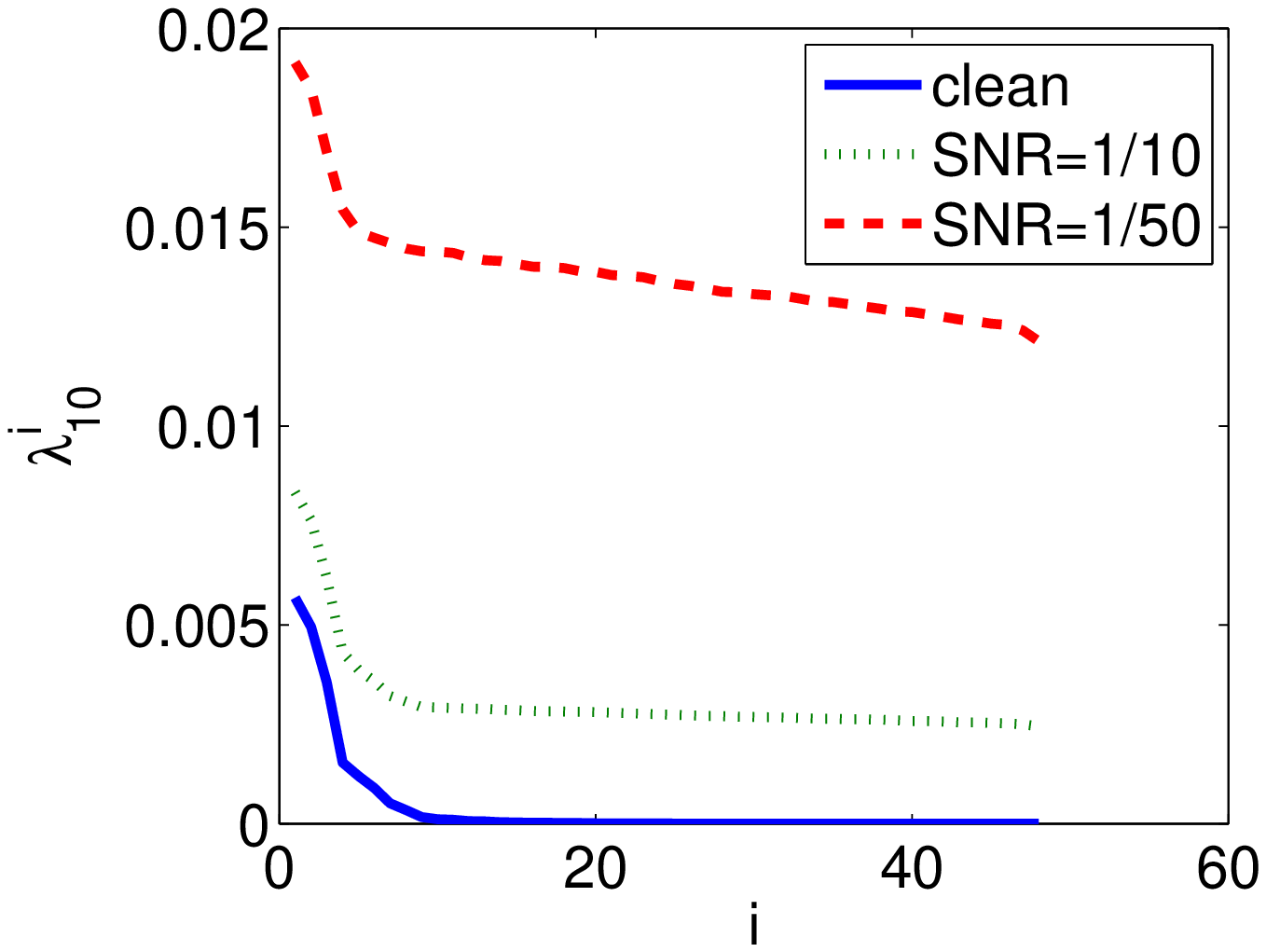}%
\label{fig:eigVal_k10}
}
\end{center}
\caption{Eigenvalues for $C^{(0)}$ and $C^{(10)}$ for simulated projection images with different signal to noise ratios.}
\label{fig:EigVal}
\end{figure}

We compared the eigenvalues in the previous experiment of the ribosome projection images with different SNRs (Fig. \ref{fig:EigVal}). As the noise variance ($\sigma^2$) increases, the number of signal components that we are able to discriminate decreases. We use the method in~\cite{Kritchman} to automatically estimate the noise variance and the number of components beyond the noise level. With the estimated noise variance $\hat{\sigma}^2$, components with eigenvalues $\lambda^i_k > \hat{\sigma}^2(1+\sqrt{1+\gamma_k^2})^2$, where $\gamma_0=\frac{p_0}{n}$ and $\gamma_k=\frac{p_k}{2n}$ for $k>0$, are chosen (the case $k=0$ is special because the expansion coefficients $a_{0,q}$ are purely real, since $a_{k,q} = (-1)^k a_{-k,q}^*$).

\begin{figure}[h!]
\begin{center}
\subfloat[]{
\includegraphics[width=0.3\columnwidth]{clean.eps} %
\label{fig:clean}
}
\subfloat[]{
\includegraphics[width=0.3\columnwidth]{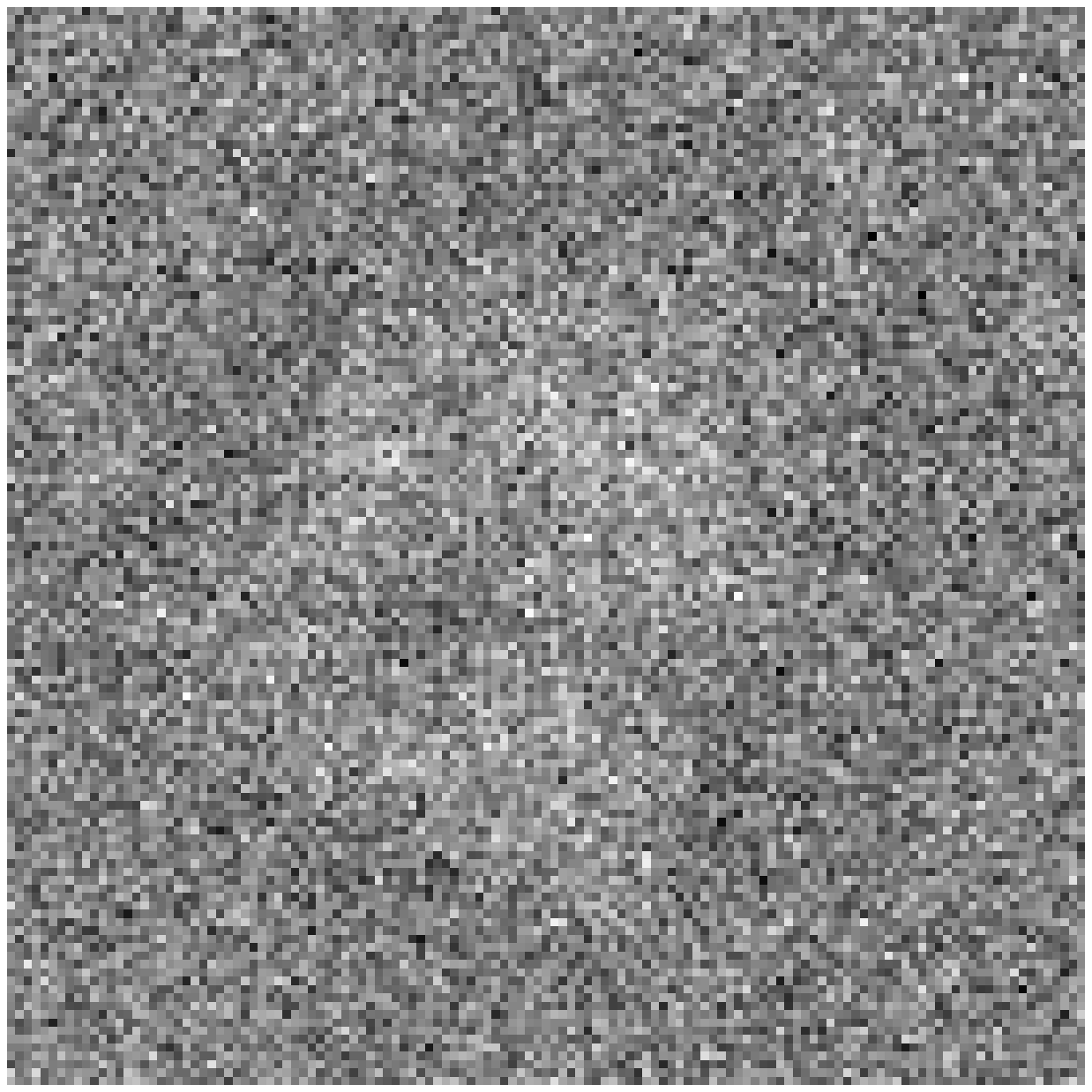} %
\label{fig:noisy}
}\\
\subfloat[]{
\includegraphics[width=0.3\columnwidth]{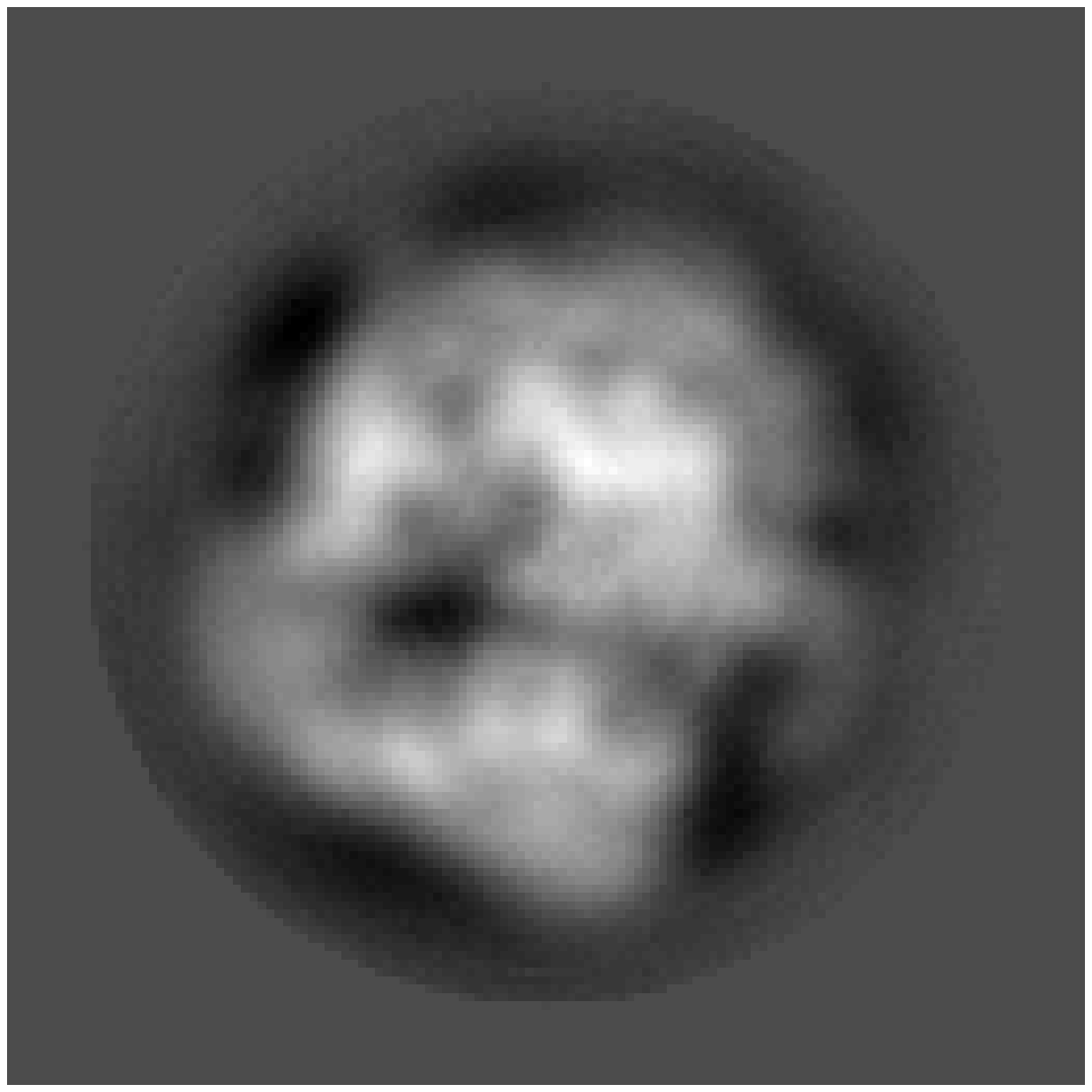} %
\label{fig:FBsPCAdenoise}
}
\subfloat[]{
\includegraphics[width=0.3\columnwidth]{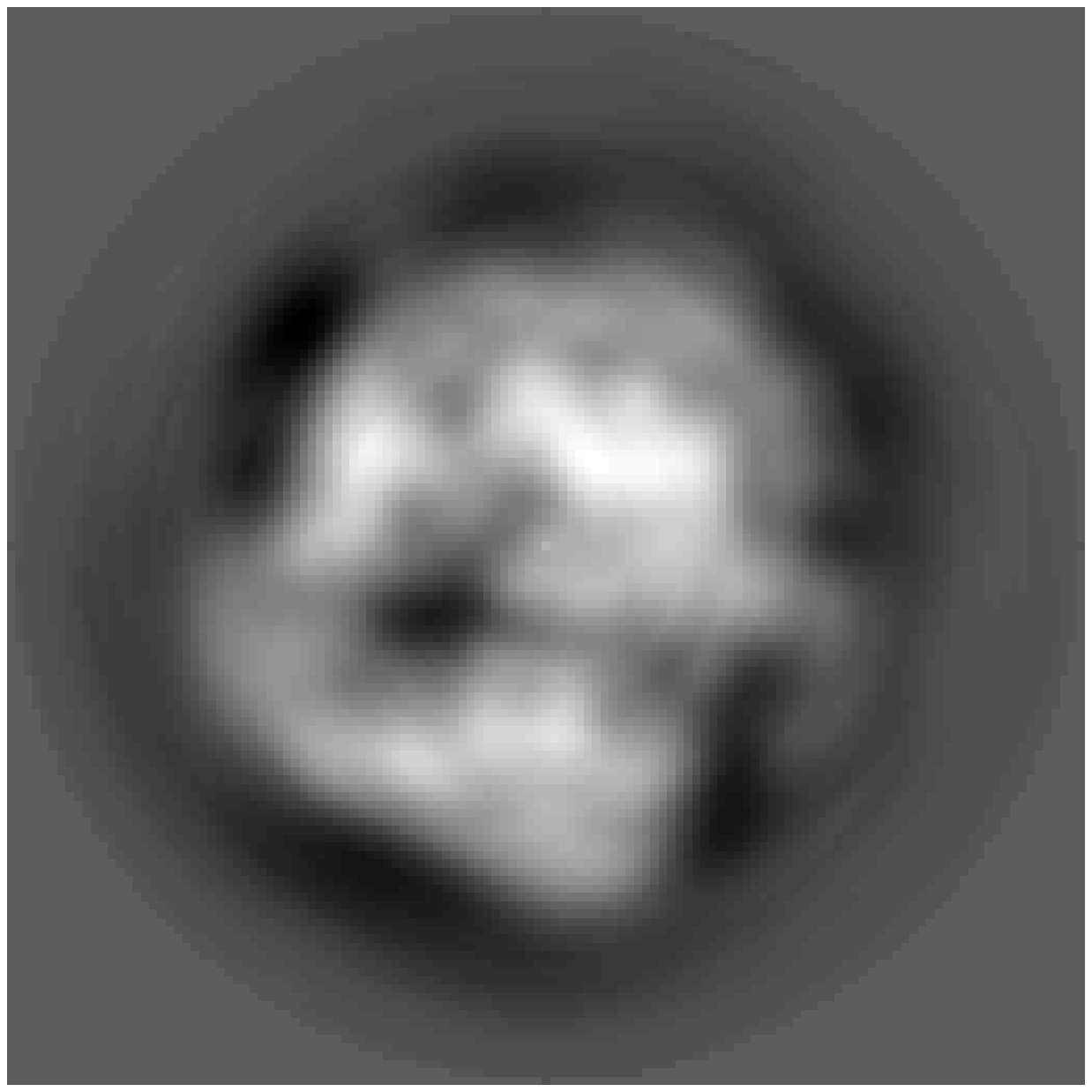} %
\label{fig:PTsPCAdenoise}
}
\subfloat[]{
\includegraphics[width=0.3\columnwidth]{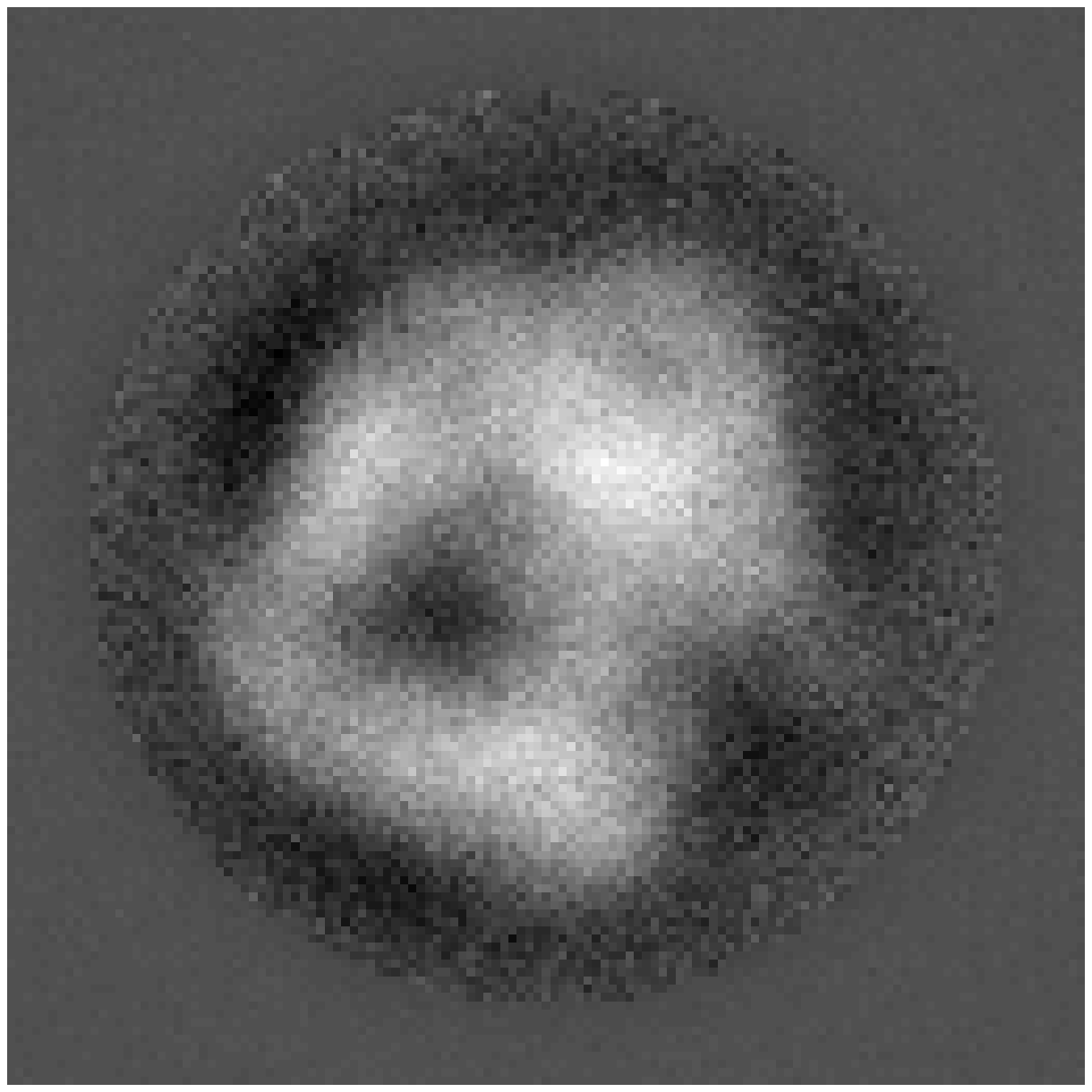} %
\label{fig:PCAdenoise}
}
\end{center}
\caption{Denoising by FBsPCA, PTsPCA and PCA. $n=10^4$ and $L=55$ pixels. (a) clean image. (b) noisy image (SNR$=\frac{1}{20}$). (c) FBsPCA denoised. (d) PTsPCA denoised. (e) PCA denoised.}
\label{fig:denoise}
\end{figure}

We compared the effects of denoising with FBsPCA, PTsPCA, and traditional PCA. The asymptotically optimal Wiener filters~\cite{Wu} designed with the eigenvalues and eigenimages from FBsPCA (PTsPCA, PCA, resp.) are applied to the noisy images. The number of signal components determined with PCA is $62$, whereas the total number of eigenimages for FBsPCA is $430$ for $10^4$ noisy images with SNR$=\frac{1}{20}$ and $L=55$ pixels. For PTsPCA, due to the non-unitary nature of the interpolation from Catersian to polar, we do not have a simple rule for choosing the number of eigenimages. Instead, we applied denoising multiple times corresponding to different number of eigenimages, and present here the denoising result with the smallest mean squared error (MSE). The optimal number of eigenimages for PTsPCA was 348. Of course, in practice the clean image is not available and such a procedure cannot be used. Still, the optimal denoising result by PTsPCA is inferior to that obtained by FBsPCA, where the number of components is chosen automatically. Even if we allow PTsPCA to use more components than FBsPCA, the denoising result does not improve. Fig. \ref{fig:denoise} shows that FBsPCA gives the best denoised image. We computed the MSE, Peak SNR (PSNR), and the structural similarity index (SSIM)~\cite{Wang} to show the effectiveness of the denoising effect using FBsPCA compared with traditional rotational variant PCA and PTsPCA (see Table~\ref{tab:denoise}).

\begin{table}
\begin{center}
\begin{tabular}{|c|c|c|c|}
\hline
 & MSE ($10^{-5}$) & PSNR(dB) & 1-SSIM ($10^{-6}$)\\
\hline
FBsPCA & $6.6$ & $20.2$  &$ 4.4 $ \\
\hline
PTsPCA & $7.7$ & $19.5$   &$ 5.9 $\\
\hline
PCA & $10.1 $ & $ 18.3 $ & $ 5.9 $ \\
\hline
\end{tabular}
\end{center}
\caption{Denoising effect: FBsPCA, PTsPCA and PCA.}
\label{tab:denoise}
\end{table}

\section{Summary and Discussion}
In this paper we adapted a sampling criterion that was originally proposed in \cite{Klug} into the framework of steerable PCA. The Fourier-Bessel transform with the modified cut-off criterion $R_{kq}\leq \pi L$ is approximately unitary and keeps the statistics of white noise approximately unchanged. This sampling criterion obtains maximum information from a set of images and their rotated and reflected copies while preventing aliasing. Instead of constructing the invariant covariance matrix from the original images and their rotated and reflected copies, we compute the covariance matrix from the truncated Fourier-Bessel expansion coefficients. The covariance matrix has a special block diagonal structure that allows us to perform PCA on different angular frequencies separately. The block diagonal structure was also observed and utilized in previous algorithms for steerable PCA. However, we show here that the block size must shrink as the angular frequency increases in order to avoid aliasing and spurious eigenimages.

While steerable PCA has found applications in computer vision and pattern recognition, this work has been mostly motivated by its application to cryo-EM. Besides compression and denoising of the experimental images, it is worth mentioning that under the assumption that the viewing directions of the images are uniformly distributed over the sphere, Kam has previously demonstrated \cite{Kam80} that the covariance matrix of the images is related to the expansion coefficients of the molecule in spherical harmonics. Our Fourier-Bessel steerable PCA can therefore be applied in conjunction to Kam's approach.

When applying PCA to cryo-EM images, one has to take into account the fact that cryo-EM images are not perfectly centered. It is well known that the unknown translational shifts can be estimated experimentally using the double exposure technique. Specifically, the images that are analyzed by PCA are obtained from an initial low-dose exposure, while the centers of the images are determined from a second exposure with higher dose. Furthermore, image acquisition is typically followed by an iterative global translational alignment procedure in which images are translationally aligned with their sample mean, that was noted earlier to be radially symmetric. It has been observed that such translational alignment measures produce images whose centers are off by only a few pixels from their true centers. Such small translational misalignments mainly impact the high frequencies of the eigenimages. The general problem of performing PCA for a set of images and their rotations and translations, or other non-compact groups of transformations is beyond the scope of this paper.

Another important consideration when applying PCA to cryo-EM images is the contrast transfer function (CTF) of the microscope. The images are not simply projections of the molecule, but are rather convolved with a CTF. The Fourier transform of the CTF is real valued and oscillates between positive and negative values. The frequencies at which the CTF vanishes are known as ``zero crossings". The images do not carry any information about the molecule at zero crossing frequencies. The CTF depends on the defocus value, and changing the defocus value changes the location of the zero crossings. Therefore, projection images from several different defocus values are acquired. When performing PCA for the images, one must take into account the fact that they belong to several defocus groups characterized by different CTFs. One possible way of circumventing this issue is to apply CTF correction to the images prior to PCA. A popular CTF correction procedure is phase-flipping which has the advantage of not altering the noise statistics. Alternatively, one can estimate the sample covariance matrix corresponding to each defocus group separately, and then combine these estimators using a least squares procedure in order to estimate the covariance matrix of projection images that are unaffected by CTF. The optimal procedure for performing PCA for images that belong to different defocus groups is a topic for future research.

Finally, we remark that the Fourier-Bessel basis can be replaced in our framework with other suitable bases. For example, the 2D prolate spheroidal wave functions (PSWF) on a disk \cite{Slepian} enjoy from many properties that make them attractive for steerable PCA. In particular, among all bandlimited functions of a given bandlimit they are the mostly spatially concentrated in the disk. They also have a separation of variables form which makes them convenient for steerable PCA. However, an accurate numerical evaluation of the PSWF requires more sophistication compared to the Bessel functions, which is the reason why we have not applied PSWFs here.

\section*{Acknowledgement}
The project described was partially supported by Award Number
R01GM090200 from the National Institute of General Medical Sciences, Award Number DMS-0914892 from the NSF, Award Numbers FA9550-09-1-0551 and FA9550-12-1-0317 from AFOSR, the Alfred P. Sloan Foundation, and Award Number LTR DTD 06-05-2012 from the Simons Foundation.


\begin{thebibliography}{99}

\bibitem{Frank} J. Frank, {\em Three-Dimensional Electron Microscopy of Macromolecular Assemblies: Visualization of Biological Molecules in Their Native State},
Oxford (2006).
\bibitem{vanHeel} M. van Heel, J. Frank, ``Use of multivariate statistics in analysing the images of biological macromolecules,'' Ultramicroscopy \textbf{6}(2) 187-194 (1981).
\bibitem{Hilai} R. Hilai and J. Rubinstein, ``Recognition of rotated images by invariant Karhunen-Lo{\'e}ve expansion,'' J. Opt. Soc. Am. A. \textbf{11,}(5) 1610-1618 (1994).
\bibitem{Perona}P. Perona, ``Deformable kernels for early vision, '' IEEE Trans. Pattern Anal. Mach. Intell. \textbf{17}(5) 488-499 (1995).
\bibitem{Uenohara} M. Uenohara and T. Kanade, ``Optimal approxmation of uniformly rotated images: Relationship between Karhunen-Lo{\'e}ve expansion and discrete cosine transform,'' IEEE Trans. Image Proces. \textbf{7}(1) 116-119 (1998).
\bibitem{Jogan} M. Jogan, E. Zagar, and A. Leonardis, ``Karhunen-Lo{\'e}ve expansion of a set of rotated templates,'' IEEE Trans. Image Process. \textbf{12}(7) 817-825 (2003).
\bibitem{Ponce} C. Ponce and A. Singer, ``Computing steerable principal components of a large set of images and their rotations,'' IEEE Transactions on Image Processing. \textbf{20,}(11) 3051-3062 (2011).
\bibitem{Klug} A. Klug and R. A. Crowther, ``Three-dimensional image reconstruction from the viewpoint of information theory,'' Nature. \textbf{238,} 435-440 (1972).
\bibitem{McMahon} J. McMahon, ``On the roots of the Bessel and certain related functions,'' Annals of Mathematics. \textbf{9,} (1) 23-30 (1894 - 1895).
\bibitem{Kritchman} S. Kritchman and B. Nadler, ``Determining the number of components in a factor model from limited noisy data,'' Chemometrics and Intelligent Laboratory Systems \textbf{94} 19–32 (2008).
\bibitem{Wang} Z. Wang, A. C. Bovik, H. R. Sheikh and E. P. Simoncelli, ``Image quality assessment: From error visibility to structural similarity,'' IEEE Transactions on Image Processing, \textbf{13}(4) 600-612 (2004).
\bibitem{Wu}A.Singer and H.-T. Wu, ``Two-dimensional tomography for noisy projections taken at unknown random directions,'' SIAM Journal on Imaging Sciences, accepted for publication. Available at    \href{https://web.math.princeton.edu/~amits/publications/random-proj_final.pdf}{this link}.
%\bibitem{Kam77} Z. Kam, ``Determination of macromolecular structure in solution by spatial correlation of scattering fluctuations,'' Macromolecules \textbf{10}(5) 927-934 (1977)
\bibitem{Kam80} Z. Kam, ``The Reconstruction of structure from electron micrographs of randomly oriented particles,'' J. theor. Biol. \textbf{82} 15-39 (1980).
\bibitem{Slepian} D. Slepian, ``Prolate spheroidal wave functions, Fourier analysis, and uncertainty
- {IV}: extensions to many dimensions, generalized prolate spheroidal wave functions,'' Bell System Technical Journal \textbf{43} 3009–3057 (1964).
\end{thebibliography}
\end{document}